\documentclass[11pt,a4paper]{article}
\usepackage{times,latexsym}
\usepackage{url}
\usepackage[T1]{fontenc}

\usepackage{amsfonts}
\usepackage{multirow}
\usepackage{colortbl}
\usepackage{pgfplots}
\usepackage{tikz}
\usepackage{pgfplotstable}
\usepackage{bbding}
\usepackage{framed}
\definecolor{shadecolor}{rgb}{0.92,0.92,0.92}
\usepackage{tcolorbox}
\usepackage{amssymb}
\usepackage{pifont}
\usepackage{setspace}
\usepackage{makecell}
\usepackage{calligra}
\usepackage{algpseudocode} 
\usepackage{subfigure}

\usepackage[ruled,linesnumbered]{algorithm2e}
\usepackage{hyperref}
\usepackage{colortbl}

\usepackage{arydshln}
\usepackage{tabu}
\usepackage{array}

\usepackage{wrapfig}
\usepackage{caption}
\usepackage{tabularx}
\usepackage{amsmath}
\usepackage{booktabs} 

\usepackage[acceptedWithA]{tacl2021v1}
\usepackage{xspace,mfirstuc,tabulary}

\newif\iftaclinstructions
\taclinstructionsfalse 
\iftaclinstructions
\renewcommand{\confidential}{}
\renewcommand{\anonsubtext}{(No author info supplied here, for consistency with
TACL-submission anonymization requirements)}
\newcommand{\instr}
\fi

\iftaclpubformat 

\else

\fi

\title{Dynamic Demonstrations Controller for In-Context Learning}

\author{Fei Zhao\quad Taotian Pang\quad Zhen Wu\quad Zheng Ma\quad \textbf{Shujian Huang}\quad \textbf{Xinyu Dai}\\
National Key Laboratory for Novel Software Technology, Nanjing University, China\\
  {\tt \{zhaof, pangtt, maz\}@smail.nju.edu.cn, }\\
  {\tt \{wuz, huangsj, daixinyu\}@nju.edu.cn}
  \\}

\date{}

\begin{document}
\maketitle
\begin{abstract}
In-context learning (ICL) is a new paradigm for natural language processing (NLP), where a large language model (LLM) observes a small number of demonstrations and a test instance as its input, and directly makes predictions without updating model parameters. Previous studies have revealed that ICL is sensitive to the selection and the ordering of demonstrations. However, there are few studies regarding the impact of the demonstration number on the ICL performance within a limited input length of LLM, because it is commonly believed that the number of demonstrations is positively correlated with model performance. In this paper, we found this conclusion does not always hold true. Through pilot experiments, we discover that increasing the number of demonstrations does not necessarily lead to improved performance. Building upon this insight, we propose a \textit{\textbf{D}ynamic \textbf{D}emonstrations \textbf{Controller}} (\textit{\textbf{D$^2$Controller}}), which can improve the ICL performance by adjusting the number of demonstrations dynamically. The experimental results show that D$^2$Controller yields a 4.6\% relative improvement on ten different sizes of LLMs across ten datasets. Besides, we also extend our method to previous ICL models and achieve competitive results.
\end{abstract}

\section{General instructions}

\begin{figure*}[t]
\centering
    \includegraphics[width=0.95\textwidth]{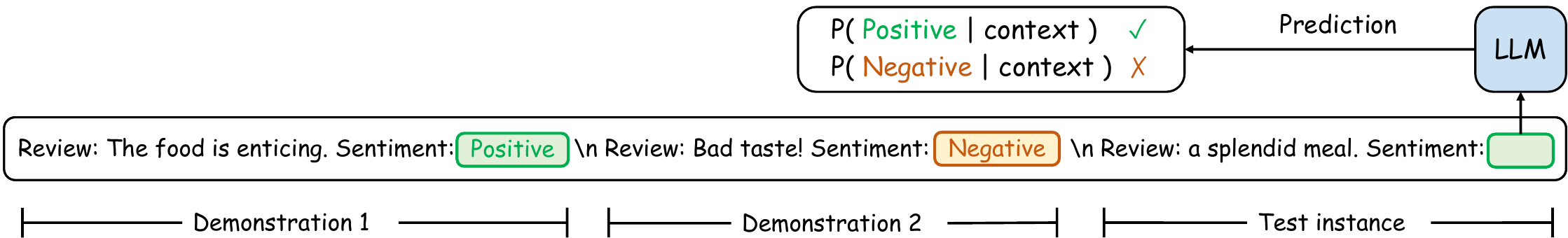}
  \caption{An example of In-Context Learning. ICL takes a small number of demonstrations and a test instance as input, with a large language model responsible for making predictions.}
  \label{fig:ICL_example}
\end{figure*}

In-context learning (ICL) is a new paradigm for performing various NLP tasks using large language models (LLMs)~\citep{DBLP:conf/nips/BrownMRSKDNSSAA20}. In ICL, by conditioning on a small number of \textit{demonstrations}, LLMs can generate predictions for a given test input without updating model parameters. Restricted by the maximum input length of LLMs, it is common to sample a small set of examples from the training dataset randomly to formulate demonstrations. Figure~\ref{fig:ICL_example} shows an example of sentiment analysis using ICL.

To improve the performance of ICL, existing work primarily focuses on designing \textit{Demonstration Selection} methods~\citep{DBLP:conf/acl-deelio/LiuSZDCC22,DBLP:conf/naacl/RubinHB22,DBLP:conf/emnlp/ZhangFT22,DBLP:journals/corr/abs-2206-08082,DBLP:journals/corr/abs-2212-04037,DBLP:conf/acl/SorensenRRSRDKF22,DBLP:journals/corr/abs-2301-11916,DBLP:journals/corr/abs-2305-04320,DBLP:journals/corr/abs-2302-13539,DBLP:conf/icml/Ye0F0K23} or finding an appropriate \textit{Demonstration Ordering}~\citep{DBLP:conf/acl/LuBM0S22,wu2022self}, since a lot of studies have revealed that ICL is sensitive to the selection as well as the ordering of demonstrations~\citep{DBLP:conf/acl-deelio/LiuSZDCC22,DBLP:conf/naacl/RubinHB22,DBLP:conf/emnlp/ZhangFT22,DBLP:conf/acl/LuBM0S22,wu2022self,DBLP:journals/corr/abs-2305-04320,DBLP:journals/corr/abs-2302-13539,dong2022survey,DBLP:conf/icml/Ye0F0K23}.

However, to the best of our knowledge, there are few studies available regarding the impact of the \textit{Demonstration Number} on the ICL performance. This scarcity may be attributed to the prevailing belief that the relation between the number of demonstrations and model performance follows a power law – as the number of demonstrations increases, model performance continues to improve~\citep{xie2022an,xu2023knn}. Nevertheless, through pilot experiments, we find this conclusion does not always hold true. Specifically, within the constraints of input length in LLMs, we systematically evaluate model performance across a spectrum ranging from the minimum to the maximum number of demonstrations. This comprehensive assessment involves five different datasets and encompasses five sizes of LLMs~\citep{DBLP:conf/nips/BrownMRSKDNSSAA20,zhang2022opt,DBLP:journals/corr/abs-2304-03208}. Our findings reveal that:
\begin{itemize}
    \item As more demonstrations are incorporated into the model input, the changes of the performance across different datasets on the same model tend to be inconsistent, with some datasets showing improvements while others experiencing declines. Similarly, the performance of different models on the same dataset also rises or falls. This suggests that increasing the number of demonstrations does not necessarily improve performance.
    \item During the transition from minimum to maximum number of demonstrations, the number of demonstrations needed for the same model to attain the optimal performance varies across different datasets. Likewise, different models exhibit variations in the number of demonstrations required to reach the optimal performance on the same dataset. This suggests that the optimal number of demonstrations may differ depending on the specific dataset and model combination.
\end{itemize}

Based on the above observation, we can infer that it is necessary to dynamically select an appropriate demonstration number for different datasets and models. Doing so not only boosts ICL performance but also can help save time and space during the inference of LLMs. To achieve this goal, we propose a \textit{\textbf{D}ynamic \textbf{D}emonstrations \textbf{Controller}} (\textit{\textbf{D$^2$Controller}}), the core idea of which involves comparing the prediction accuracy of different demonstration numbers on a small set of specially selected evaluation examples. The key challenge of this idea is determining which evaluation examples should be chosen to provide a correct assessment for different demonstration numbers. To tackle this challenge, we design a metric named \textit{\textbf{I}ntra-\textbf{I}nter-\textbf{C}lass \textbf{Score}} (\textit{\textbf{IICScore}}) to guide the D$^2$Controller to select suitable evaluation examples from the training dataset. Finally, we apply D$^2$Controller to different sizes of LLMs and obtain a 4.6\% relative improvement over ten datasets. Besides, we extend our method to previous ICL models and achieve competitive results.

Our contributions are summarized as follows: (1) We comprehensively analyze the effects of the number of demonstrations on ICL performance under a limited input length of LLM, and find that the number of demonstrations may not necessarily be positively correlated with model performance; (2) We propose a method named D$^2$Controller, which not only boosts ICL performance but also saves time and space during inference of the LLMs; (3) We apply our method to ten different sizes of LLMs and realize an average of 4.6\% relative improvement across ten datasets. Moreover, we also extend our method to previous ICL models and yield competitive results.

\begin{figure*}[t]
\centering
\begin{minipage}{0.32\linewidth}
    \centering
    \includegraphics[width=\linewidth]{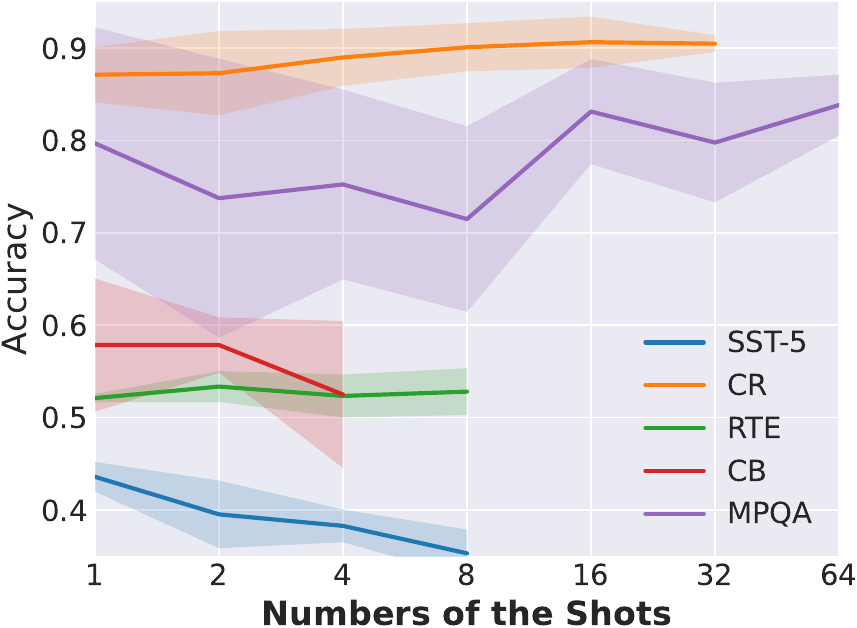}
    \caption{The influence of varying the number of demonstrations on the Cerebras-GPT-6.7B model across five different datasets.}
    \label{fig:pilot_6.7B}
\end{minipage}
\hfill
\begin{minipage}{0.32\linewidth}
    \centering
    \includegraphics[width=\linewidth]{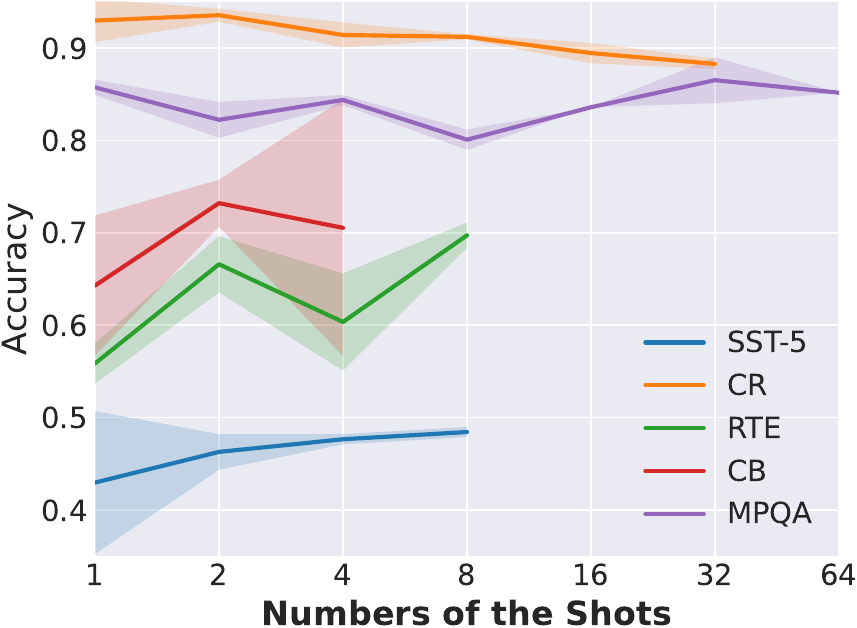}
    \caption{The effect of varying the number of demonstrations on the result of the GPT-3-175B model across five different datasets.}
    \label{fig:pilot_175B}
\end{minipage}
\hfill
\begin{minipage}{0.32\linewidth}
    \centering
    \includegraphics[width=\linewidth]{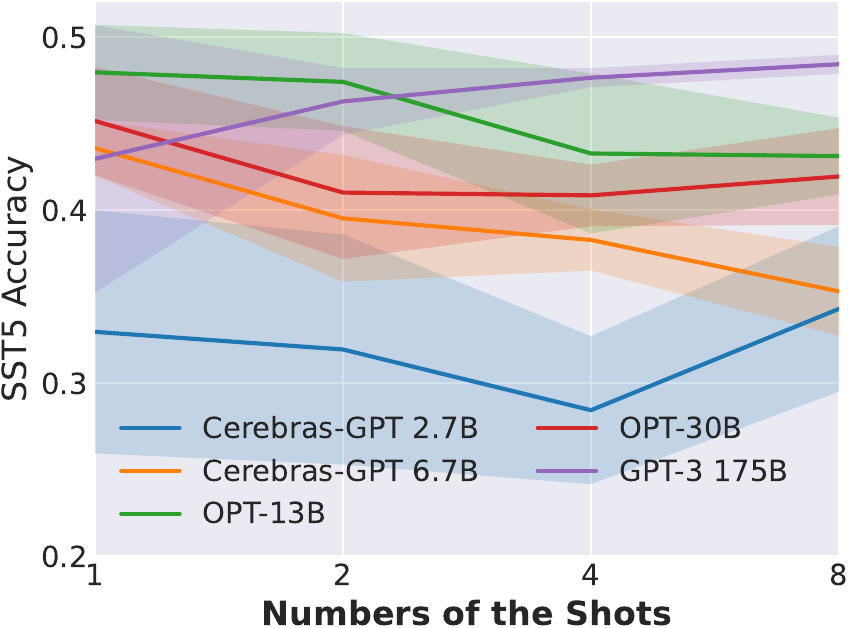}
    \caption{The Accuracy of five different sizes of LLMs on the SST-5 dataset under varying the number of demonstrations.}
    \label{fig:pilot_sst5}
\end{minipage}
\end{figure*}

\section{Background}\label{seq:bg}
In this section, we review the definition of In-Context Learning and the $k$-shot setting. 

\paragraph{Notation} We use $\boldsymbol{\theta}$ to denote an LLM. The training dataset is denoted as $\mathcal{D}$. A training example $(x_i,y_i)$ consists of a sentence $x_i$ and a label $y_i$. The sentence of a training example is also referred to as an \textit{instance}. We use $\mathcal{I}_{\mathcal{D}}=\{x_i\}_{i=1}^{|\mathcal{D}|}$ to represent all instances of training examples in $\mathcal{D}$. The label space is denoted as $\mathcal{Y}$. In this paper, we focus on ICL for text classification tasks. Each training example belongs to a certain class. The set of classes is represented as $\mathcal{C}$ and a class $c\in\mathcal{C}$ has a one-to-one correspondence with a label $y^c\in\mathcal{Y}$, \textit{i.e.}, $|\mathcal{Y}|=|\mathcal{C}|$. For example, the label ``not entailment'' corresponds to the class in which premise sentences do not entail hypothesis sentences. 

\subsection{In-Context Learning}\label{In_Context_Learning}
Given an LLM $\boldsymbol{\theta}$, a group of $n$ in-context examples $\{x_i, y_i\}_{i=1}^n$ sampled from training dataset $\mathcal{D}$ ($n\ll|\mathcal{D}|$), and a test instance $x_{\text{test}}$, ICL first formulates in-context examples in the format of the input-label pairs which are named the \textit{demonstrations} (See Appendix~\ref{sec:Template} for details) via templates, and then concatenates them together along with a test input to construct a prompt $P$:
\begin{equation}\label{eq:form_prompt}
P=\Omega(x_1, y_1) \oplus \cdots \oplus \Omega(x_n, y_n) \oplus \Omega(x_\text{test}, *),
\end{equation}
where $\Omega(\cdot,\cdot)$ denotes template-based transformation and $\oplus$ means concatenation operation. Notice that there is a verbalization process $\pi(\cdot)$ inside $\Omega(\cdot,\cdot)$, which maps the label $y_i$ to a token $v_i$ in the LLM vocabulary. The $y_i$ and $v_i$ can be different. For example, the label ``not entailment'' can be mapped to the token ``false''. We denote the mapping token space as $\mathcal{V}$ and we have $|\mathcal{Y}|=|\mathcal{V}|$ (See Appendix~\ref{sec:Template} for details). Finally, The prompt $P$ is fed into the LLM $\boldsymbol{\theta}$ to predict the label of the test instance $x_\text{test}$:
\begin{equation}
\hat{y}_{\text{test}}=\pi^{-1}(\underset{v \in \mathcal{V}}{\operatorname{\arg\max}}\quad \boldsymbol{p}(v|P,\boldsymbol{\theta})),
\end{equation}
where $\pi(\cdot)^{-1}$ denotes the inverse mapping from the token $v_i$ to the label $y_i$.

\subsection{$k$-shot Setting} 
For text classification tasks, each prompt $P$ is formulated in the class balance way, \textit{i.e.}, the demonstrations of each class are contained in a prompt $P$ and the numbers of them are the same\footnote{For example, in a 2-class sentiment analysis task, a prompt $P$ contains demonstrations from both the positive sentiment class and the negative sentiment class.}. Among them, the number of demonstrations of each class is also called the \textit{shot number}, denoted as $k$. Based on this, the $k$-shot setting means a prompt $P$ contains $k$ demonstrations for each class. In other words, the total demonstration number $n$ of each prompt $P$ is equal to $k|\mathcal{C}|$. In this paper, we vary the number of demonstrations $n$ by changing the $k$-shot setting.


Due to the input length limitation of LLMs, there exists a maximum $k$, denoted as $k_{\max}$, for every dataset. All feasible choices of $k$ for a dataset form a set $\mathcal{K}=\{1,2,\cdots,k_{\max}\}$ (Appendix~\ref{sec:Datasets} provides the calculation method for $k_{\max}$ and the value of $k_{\max}$ for each dataset).

\section{Pilot Experiments}\label{sec:Pilot_Experiments}
In this section, we conduct pilot studies to answer the following research question: \textit{Does model performance consistently improve when more demonstrations are added to prompts}?

\subsection{Experimental Setup} 
We conduct pilot experiments across five text classification datasets on six different sizes of LLMs, including two Cerebras-GPT models~\citep{DBLP:journals/corr/abs-2304-03208} (with 2.7B and 6.7B parameters), two OPT models~\citep{zhang2022opt} (with 13B and 30B parameters), a GPT-3 model~\citep{DBLP:conf/nips/BrownMRSKDNSSAA20} (with 175B parameters) and a GPT-4 model~\citep{achiam2023gpt}. We adopt \textit{Accuracy} as the evaluation metric for model performance~\citep{DBLP:conf/acl/LuBM0S22,DBLP:conf/emnlp/ZhangFT22}.  Following~\citep{DBLP:conf/acl/LuBM0S22,xu2023knn}, we randomly sample $256$ examples from the validation set for each dataset to evaluate the accuracy and report the average performance and standard deviation based on $5$ different seeds.

For each dataset, we iteratively test the model performance from $1$-shot setting to $k_{\max}$-shot setting on five sizes of LLMs. Figure~\ref{fig:pilot_6.7B} and Figure~\ref{fig:pilot_175B} show the performance curves of five datasets on the Cerebras-GPT 6.7B model and the GPT-3 175B model, respectively. Figure \ref{fig:pilot_sst5} shows performance curves of the SST5 dataset on the five different sizes of LLMs. More results can be found in Appendix~\ref{Additional_Results}.



\subsection{Analysis}


Based on these results, we conducted the following analysis:

\paragraph{Increasing the number of demonstrations does not necessarily improve the model performance.} In Figure \ref{fig:pilot_6.7B}, we can see that when more demonstrations are added to prompts, \textit{i.e.}, the shot number is increased, the model performance goes up or down on five different datasets. From a local point of view, when changing from an $8$-shot setting to a $16$-shot setting on the MPQA dataset, the model performance increases from $71.5$ to $83.1$, while the accuracy drops to $79.8$ with a $32$-shot setting. Likewise, on the CB dataset, the accuracy declines when shifting from a $2$-shot setting to a $4$-shot setting. Furthermore, when providing more demonstrations on the SST-5 dataset, the model's performance consistently decreases. From the perspective of a general trend, the accuracy improves on the MPQA dataset while declines on the CB and SST-5 datasets. Similar observations can be found in the results of the GPT-3 175B model, shown in Figure \ref{fig:pilot_175B}. Besides, the performance of different models on the same dataset also rises or falls. As shown in Figure \ref{fig:pilot_sst5}, when changing from a $1$-shot setting to a $8$-shot setting, the accuracy of the SST5 dataset on the OPT-13B model continues to decrease, while that on the GPT-3-175B model keeps rising. These observations indicate that the inclusion of more demonstrations does not guarantee improved performance.




\begin{figure*}[h]
\centering
\includegraphics[width=0.95\textwidth]{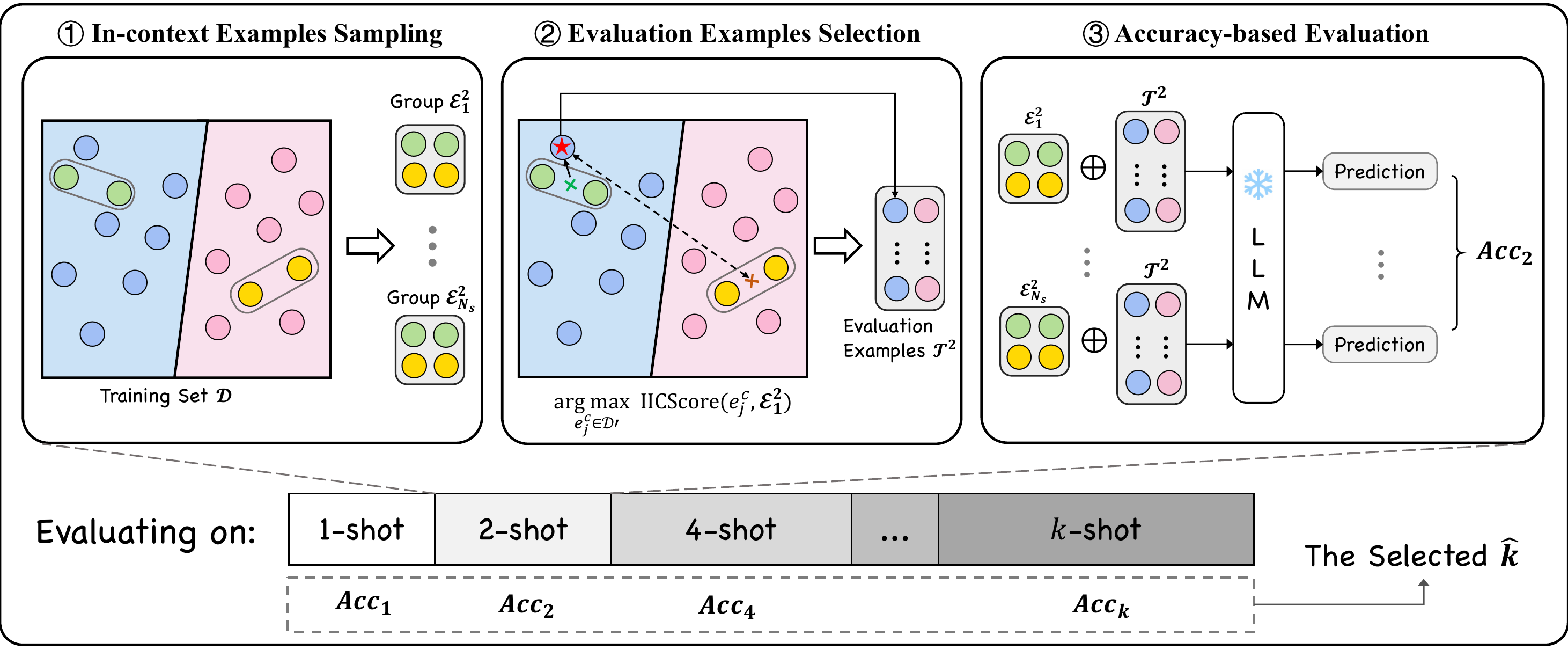}
\caption{The whole process of the D$^2$Controller on a $2$-class classification task.}
\label{fig:st-sel}
\end{figure*}

\paragraph{The optimal $k$-shot setting differs depending on specific datasets and models.} Here we define the $k$-shot setting under which a dataset acquires the highest accuracy as the optimal $k$-shot setting. From Figure~\ref{fig:pilot_175B}, we can tell that the optimal $k$-shot setting for each dataset is different: $2$-shot setting for CR and CB datasets, $8$-shot setting for RTE and SST5 dataset and $32$-shot setting for MPAQ dataset. Jointly observing Figure~\ref{fig:pilot_6.7B} and Figure~\ref{fig:pilot_175B}, we find that the optimal $k$-shot settings for the same dataset on different models can be different. The curves in Figure \ref{fig:pilot_sst5} further support this finding.

From the above analysis, we can infer that to achieve better performance in ICL, it is not appropriate to simply use the $k_{\max}$-shot setting for each dataset or the same $k$-shot setting for all datasets. The latter is a strategy widely adopted in previous work~\citep{DBLP:conf/acl/LuBM0S22,xu2023knn}. Instead, we should dynamically decide $k$-shot settings for ICL depending on specific datasets and models. 

Finally, we analyze the reasons behind these phenomena. Specifically, we speculate that adding a demonstration to a prompt will have two effects: (1) providing more information to the prompt, resulting in improvement in performance; (2) increasing the length of the prompt, which causes the distribution of the input to become more different from that of the pre-training corpus of LLMs, leading to difficulty in understanding the prompt and reducing performance. When more demonstrations are added, the direction of the change in performance depends on which effect is more influential. For different datasets and LLMs, when adding more demonstrations, the strengths of Effect (1) and Effect (2) are different, leading to the variation observed in pilot experiments and also causing the difference in the optimal $k$.

\section{Methodology}
Based on the observations of the pilot study, we propose a \textit{\textbf{D}ynamic \textbf{D}emonstrations \textbf{Controller}} (\textit{\textbf{D$^2$Controller}}), which dynamically finds a suitable $k$ from the feasible shot numbers set $\mathcal{K}$ for each dataset. An intuitive way to decide an appropriate $k$ for a specific dataset is to compare the average prediction accuracy of different $k$-shot settings on a set of evaluation examples and make a choice. The key challenge of such idea lies in that on which evaluation examples we can obtain the proper evaluation for each $k$-shot setting.

To tackle the above challenge, we propose a metric named \textit{\textbf{I}ntra-\textbf{I}nter-\textbf{C}lass \textbf{Score}} (\textit{\textbf{IICScore}}) to guide us to choose the representative evaluation examples for each group of in-context examples from the training dataset. The whole process to evaluate each $k$-shot setting is divided into three steps: (1) In-context examples sampling. (2) IICScore-guided evaluation examples selection. (3) Accuracy-based evaluation. The workflow of D$^2$Controller is illustrated in Figure \ref{fig:st-sel}. 

\subsection{In-context Examples Sampling}
In the first step, we sample $N_s$ groups of in-context examples for each $k$-shot setting, which are evaluated later. A group of in-context examples is denoted as:
\begin{equation}
\mathcal{E}^k_i=\{(x_{ij},y_{ij})|j=1,\cdots ,k|\mathcal{C}|\},i=1,\cdots,N_s,
\end{equation}
where $k$ denotes the $k$-shot setting. All in-context examples are removed from the training set $\mathcal{D}$ and the remaining ones formulate the candidate set $\mathcal{D}'$, from which we select evaluation examples in the next step. 

\subsection{IICScore-guided Evaluation Examples Selection}\label{SL_Selection}
In this step, we aim to select a set of examples from $\mathcal{D}'$ to properly evaluate the performance of each group of in-context examples. By synthesizing their performance we can further obtain the assessment of each $k$-shot setting. 

For each group of in-context examples, to fully evaluate their ability, we select their similar and dissimilar examples from $\mathcal{D}'$ as representative evaluation examples. The idea behind the selection is: (1) a group of in-context examples should be able to guide LLMs to correctly predict on examples that are similar to them; (2) they should also have ability to guide LLMs to make correct predictions on some of different examples to them. By evaluating on these two types of examples, we can obtain a comprehensive assessment of performance of each group of in-context examples.




To measure similarities, we first input each sentence $x$ into LLMs and obtain its vector representation $\boldsymbol{x}$. Then, when searching similar examples for class-$c$ in-context examples, we expect them to be not only close to the in-context examples of class $c$, but also far from those of other classes. To this end, we propose IICScore, which considers both intra-class distance and inter-class distance, to guide our selection process. IICScore is defined as: 
\begin{align}\label{eq:IICScore}
    \text{IICScore}(e^c_j, \mathcal{E}^k_i)=-\text{KL}(\boldsymbol{x}_j^c,\Bar{\boldsymbol{x}}^c_{\mathcal{I}_{\mathcal{E}^k_i}})\\\nonumber
    +\sum_{c'\in\mathcal{C},c'\neq c}\frac{|\mathcal{D}'^{c'}|}{|\mathcal{D}'|}\text{KL}(\boldsymbol{x}_j^c,\Bar{\boldsymbol{x}}^{c'}_{\mathcal{I}_{\mathcal{E}^k_i}}),
\end{align}
where $e^c_j=(x^c_j,y^c)\in\mathcal{D}'$ is a candidate example of class $c$, $\boldsymbol{x}^c_j$ denotes the vector representation of instance $x^c_j$, $\mathcal{I}_{\mathcal{E}^k_i}$ denotes the set of all instances in $\mathcal{E}^k_i$,  $\Bar{\boldsymbol{x}}^c_{\mathcal{I}_{\mathcal{E}^k_i}}$ is the average representation of class-$c$ instances in $\mathcal{I}_{\mathcal{E}^k_i}$, $\mathcal{D}'^{c'}$ means the set of class-$c'$ candidate examples, and $\text{KL}(\cdot,\cdot)$ is the KL divergence. The $\frac{|\mathcal{D}'^{c'}|}{|\mathcal{D}'|}$ is a scale factor that balances the contribution of intra-class distance and inter-class distance. Given that the $\boldsymbol{x}^c_j$ is a distribution, we choose KL divergence to measure distances. The higher the IICScore is, the more similar that candidate example $e^c_j$ is to class-$c$ in-context examples. For each group $\mathcal{E}^k_i$, the example with the highest IICScore in each class is selected as follows:
\begin{equation}
\tilde{e}^{c}_{\mathcal{E}^k_i}=\underset{e^c_j\in\mathcal{D}'}{\operatorname{\arg\max}}\quad\text{IICScore}(e^c_j,\mathcal{E}^k_i).
\end{equation}
In total, $|\mathcal{C}|$ similar examples are selected for each $\mathcal{E}^k_i$.

For dissimilar examples, since similar examples of any two different groups $\mathcal{E}^k_i$ and $\mathcal{E}^k_j$ are different, the similar example $\tilde{e}^c_{\mathcal{E}^k_j}$ is naturally a dissimilar example for $\mathcal{E}^k_i$. Gathering all $N_s|\mathcal{C}|$ selected examples to form the set of evaluation examples $\mathcal{T}^k$, there are $|\mathcal{C}|$ similar examples and $(N_s-1)|\mathcal{C}|$ dissimilar examples for each group of in-context examples.

\subsection{Accuracy-based Evaluation}
In the last step, we iteratively combine in-context examples with every evaluation example in $\mathcal{T}^k$ to create prompts (As shown in Equation \ref{eq:form_prompt}). After that, the prompts are fed into LLMs to get predictions. The average prediction accuracy of $N_s$ group of in-context examples is treated as the performance of $k$-shot setting:
\begin{equation}
    \text{Acc}_k=\frac{1}{N_s}\sum_{i=1}^{N_s}(\frac{1}{|\mathcal{T}^k|}\sum_{j=1}^{|\mathcal{T}^k|}\mathbb{I}(\hat{y}_{j,\mathcal{E}_i^k}=y_{j})),
\end{equation}
where $\hat{y}_{j,\mathcal{E}_i^k}$ means the predicted label of $j$-th example in $\mathcal{T}^k$ using demonstrations transformed from $\mathcal{E}_i^k$ and $\mathbb{I}$ is the indicator function. After testing the performance of all feasible $k$-shot settings, we choose the one with the best performance as follows:
\begin{equation}
    \hat{k}=\underset{k\in\mathcal{K}}{\operatorname{\arg\max}}\quad\text{Acc}_k.
\end{equation}


The algorithm details of the D$^2$Controller are presented in Appendix~\ref{Algorithm_details}. It is worth mentioning that our approach is model-agnostic, allowing it to be combined with LLMs of different sizes and applied to previous ICL methods.

\begin{table*}[t]
\centering
\scalebox{0.68}{
\begin{tabular}{llccccccccccc}

\toprule
& & \textbf{SST-2} & \textbf{SST-5} & \textbf{DBPedia} & \textbf{MR} & \textbf{CR} & \textbf{MPQA} & \textbf{Subj} & \textbf{AGNews} & \textbf{RTE} & \textbf{CB} & \textbf{Average}\\

\midrule

\multirow{3}{*}{\makecell[l]{GPT-2\\0.3B}} & Default & 58.1$_{13.1}$ & 24.1$_{7.4}$ & 60.6$_{7.2}$ & 54.2$_{10.6}$ & 50.6$_{0.4}$ & 59.6$_{15.8}$ & 53.4$_{5.3}$ & 48.7$_{8.5}$ & 51.3$_{1.7}$ & 48.6$_{6.4}$ & 50.9\\

& D$^2$Controller & 74.1$_{9.3}$ & 31.6$_{8.6}$ &	60.6$_{7.2}$ & 53.8$_{7.0}$ & 67.7$_{11.4}$ & 57.1$_{9.7}$ &	53.8$_{4.2}$ & 48.7$_{8.5}$ & 48.7$_{2.9}$ & 48.6$_{6.4}$ & \textbf{54.5}\\

& Oracle & 74.1$_{9.3}$ & 31.6$_{8.6}$ & 60.6$_{7.2}$ & 56.0$_{9.9}$ & 67.7$_{11.4}$ & 64.5$_{16.0}$ & 58.6$_{12.8}$ & 49.4$_{18.4}$ & 51.3$_{1.7}$ & 50.0$_{9.2}$ & 56.4\\

\midrule
\multirow{3}{*}{\makecell[l]{GPT-2\\0.8B}} & Default  & 71.8$_{12.1}$& 37.8$_{6.8}$ & 63.4$_{6.0}$ & 71.1$_{15.6}$ & 80.5$_{11.4}$ & 65.8$_{11.3}$ & 59.9$_{12.2}$ & 65.6$_{17.2}$ & 53.1$_{3.4}$ & 37.1$_{14.5}$ & 60.6\\

& D$^2$Controller & 65.9$_{15.2}$ & 37.5$_{5.1}$ & 63.4$_{6.0}$ & 71.1$_{15.6}$ & 80.5$_{11.4}$ & 70.5$_{5.2}$ & 69.4$_{12.4}$ & 65.6$_{17.2}$ & 53.1$_{3.4}$ & 47.5$_{3.2}$ & \textbf{62.4}\\



& Oracle & 71.8$_{12.1}$ & 39.6$_{5.1}$ & 63.4$_{6.0}$ & 71.1$_{15.6}$ & 80.5$_{11.4}$ & 74.5$_{8.8}$ & 69.4$_{12.4}$ & 65.6$_{17.2}$ & 53.8$_{4.4}$ & 49.3$_{3.7}$ & 63.9\\

\midrule
\multirow{3}{*}{\makecell[l]{GPT-2\\1.5B}} & Default & 70.3$_{6.6}$ & 35.4$_{8.4}$ & 82.0$_{2.0}$ & 52.0$_{3.8}$ & 52.0$_{3.2}$ & 66.7$_{8.2}$ & 57.3$_{10.5}$ & 78.2$_{6.7}$ & 53.1$_{1.7}$ & 52.9$_{6.3}$ & 60.0\\
& D$^2$Controller & 81.3$_{5.4}$ & 35.4$_{8.4}$ & 82.0$_{2.0}$ & 72.2$_{13.9}$ & 66.2$_{16.7}$ & 83.9$_{1.5}$ & 64.1$_{11.3}$ & 78.2$_{6.7}$ & 53.1$_{2.9}$ & 52.9$_{6.3}$ & \textbf{67.0}\\



& Oracle & 81.3$_{5.4}$ & 40.6$_{5.4}$ & 82.0$_{2.0}$ & 72.2$_{13.9}$ & 66.2$_{16.7}$ & 83.9$_{1.5}$ & 64.1$_{11.3}$ & 81.3$_{7.5}$ & 53.1$_{2.9}$ & 57.9$_{9.8}$ & 68.2\\

\midrule
\multirow{3}{*}{\makecell[l]{Cerebras-GPT\\2.7B}} & Default & 65.5$_{13.8}$ & 28.4$_{4.3}$ & 81.8$_{1.4}$ & 65.1$_{11.2}$ & 85.8$_{4.2}$ & 64.2$_{11.6}$ & 69.3$_{14.4}$ & 69.5$_{3.2}$ & 48.1$_{1.1}$ & 52.5$_{9.5}$ & 63.0\\
& D$^2$Controller & 77.3$_{7.7}$ & 34.3$_{4.8}$ & 81.8$_{1.4}$ & 76.0$_{7.7}$ & 87.4$_{1.5}$ & 81.6$_{2.1}$ & 74.2$_{7.6}$ & 77.3$_{4.1}$ & 48.0$_{1.1}$ & 54.6$_{2.7}$ & \textbf{69.3}\\
 
& Oracle & 80.7$_{9.1}$ & 34.3$_{4.8}$ & 81.8$_{1.4}$ & 76.0$_{7.7}$ & 87.4$_{1.5}$ & 82.9$_{3.0}$ & 74.2$_{7.6}$ & 77.3$_{4.1}$ & 49.6$_{2.3}$ & 55.7$_{5.0}$ & 70.0\\

\midrule
\multirow{3}{*}{\makecell[l]{Cerebras-GPT\\6.7B}} & Default & 83.4$_{8.5}$ & 38.3$_{1.8}$ & 87.0$_{2.4}$ & 88.0$_{1.1}$ & 89.0$_{3.1}$ & 75.2$_{10.3}$ & 72.0$_{14.5}$ & 79.2$_{2.4}$ & 52.3$_{2.3}$ & 52.5$_{8.0}$ & 71.7\\
& D$^2$Controller & 82.0$_{11.3}$ & 39.5$_{3.7}$ & 87.0$_{2.4}$ & 86.8$_{1.9}$ & 90.5$_{0.9}$ & 83.8$_{3.3}$ & 79.2$_{12.5}$ & 80.2$_{1.5}$ & 52.8$_{2.5}$ & 57.9$_{7.2}$ & \textbf{74.0}\\  
& Oracle & 88.6$_{2.7}$ & 43.6$_{1.6}$ & 87.0$_{2.4}$ & 88.0$_{1.1}$ & 90.6$_{2.8}$ & 83.8$_{3.3}$ & 79.2$_{12.5}$ & 80.2$_{1.5}$ & 53.4$_{1.7}$ & 57.9$_{3.0}$ & 75.2\\

\midrule
\multirow{3}{*}{\makecell[l]{LLAMA\\7B}} & Default         & $92.6_{0.6}$ & $38.2_{4.5}$ & $81.2_{1.4}$ & $92.4_{0.5}$ & $92.0_{1.5}$ & $84.4_{2.9}$ & $52.0_{0.0}$ & $85.6_{1.5}$ & $74.2_{3.0}$ & $74.6_{11.3}$ & 76.7\\
& D$^2$Controller & 92.6$_{0.6}$  & 38.2$_{4.5}$  & 81.2$_{1.4}$    & 92.4$_{0.5}$   & 92.0$_{1.5}$    & 84.4$_{2.9}$    & 52.0$_{0.0}$    & 86.2$_{1.0}$    & 74.2$_{3.0}$   & 84.3$_{3.4}$  & \textbf{77.8}\\
& Oracle & 93.4$_{0.6}$ & 39.5$_{8.1}$ & 81.2$_{1.4}$ & 93.2$_{1.0}$ & 92.4$_{0.8}$ & 86.9$_{1.3}$ & 52.0$_{0.0}$ & 87.0$_{2.3}$ & 74.2$_{3.0}$ & 84.3$_{3.4}$ & 78.4\\
\midrule
\multirow{3}{*}{\makecell[l]{LLAMA-2\\7B}} & Default & 92.6$_{2.0}$ & 47.2$_{1.7}$ & 80.6$_{1.0}$ & 92.8$_{1.0}$ & 89.5$_{3.2}$ & 75.9$_{9.9}$ & 52.0$_{0.0}$ & 84.5$_{4.4}$ & 70.6$_{4.1}$ & 70.4$_{12.6}$ & 75.6\\
& D$^2$Controller & 91.7$_{3.9}$  & 49.0$_{2.4}$ & 80.6$_{1.0}$ & 93.4$_{0.6}$ & 89.1$_{2.7}$ & 84.2$_{2.7}$ & 52.0$_{0.0}$ & 84.5$_{4.5}$ & 70.6$_{4.1}$ & 68.8$_{1.3}$ & \textbf{76.4}\\
& Oracle & 93.8$_{0.6}$ & 49.0$_{2.4}$ & 80.6$_{1.0}$ & 93.4$_{0.6}$ & 89.7$_{2.4}$ & 87.0$_{1.8}$ & 52.0$_{0.0}$ & 86.4$_{0.9}$  & 72.3$_{3.9}$ & 70.4$_{12.6}$ & 77.5\\
\midrule
\multirow{3}{*}{\makecell[l]{OPT\\13B}} & Default & 81.2$_{6.7}$ & 43.3$_{4.6}$ & 92.3$_{2.1}$ & 87.8$_{2.7}$ & 91.4$_{3.3}$ & 75.0$_{6.7}$ & 79.1$_{12.7}$ & 81.9$_{2.9}$ & 54.4$_{4.2}$ & 58.9$_{8.1}$ & 74.5\\

& D$^2$Controller & 90.2$_{5.8}$ & 43.3$_{4.6}$ & 92.3$_{2.1}$ & 87.8$_{2.7}$ & 91.3$_{2.1}$ & 72.0$_{9.4}$ & 91.6$_{2.0}$ & 82.6$_{1.5}$ & 55.8$_{3.1}$ & 58.9$_{8.1}$ & \textbf{76.6}\\
& Oracle & 90.9$_{3.7}$ & 48.0$_{2.8}$ & 92.3$_{2.1}$ & 91.8$_{0.6}$ & 93.3$_{1.2}$ & 78.6$_{7.3}$ & 91.6$_{2.0}$ & 82.6$_{1.5}$ & 55.8$_{3.1}$ & 73.2$_{12.4}$ & 79.8\\

\midrule

\multirow{3}{*}{\makecell[l]{OPT\\30B}} & Default  & 92.3$_{1.3}$ & 40.9$_{1.8}$ & 91.7$_{3.7}$ & 91.8$_{2.1}$ & 87.3$_{3.3}$ & 78.8$_{6.2}$ & 76.1$_{4.9}$ & 78.7$_{3.6}$ & 63.0$_{3.1}$ & 60.0$_{8.2}$ & 76.1\\

& D$^2$Controller & 92.3$_{1.3}$ & 42.0$_{2.8}$ & 91.7$_{3.7}$ & 93.4$_{1.1}$ & 87.3$_{2.7}$ & 85.7$_{3.8}$ & 83.4$_{8.6}$ & 76.7$_{4.5}$ & 61.6$_{2.8}$ & 60.0$_{8.2}$ & \textbf{77.4}\\

& Oracle & 92.8$_{1.6}$ & 45.2$_{3.1}$ & 91.7$_{3.7}$ & 93.4$_{1.1}$ & 87.7$_{3.9}$ & 85.7$_{3.8}$ & 83.4$_{8.6}$ & 78.7$_{3.6}$ & 63.0$_{3.1}$ & 60.0$_{8.2}$ & 78.1\\

\midrule

\multirow{3}{*}{\makecell[l]{GPT-3\\175B}} & Default & 94.0$_{1.4}$ & 47.7$_{0.6}$ & 90.2$_{2.8}$ & 94.1$_{0.6}$ & 91.4$_{0.0}$ & 84.4$_{0.6}$ & 71.1$_{2.2}$ & 86.9$_{1.4}$ & 60.4$_{5.3}$ & 70.5$_{13.9}$ & 79.1\\

& D$^2$Controller & 94.0$_{1.4}$ & 48.4$_{0.6}$ & 90.2$_{2.8}$ & 95.5$_{0.8}$ & 93.0$_{2.3}$ & 84.4$_{0.6}$ & 87.3$_{4.7}$ & 86.9$_{1.4}$ & 66.6$_{3.0}$ & 73.2$_{2.5}$ & \textbf{82.0}\\

& Oracle & 94.1$_{0.0}$ & 48.4$_{0.6}$ & 90.2$_{2.8}$ & 95.5$_{0.3}$ & 93.6$_{2.8}$ & 86.5$_{2.5}$ & 87.3$_{4.7}$ & 86.9$_{1.4}$ & 69.7$_{1.4}$ & 73.2$_{2.5}$ & 82.6\\

\bottomrule
\end{tabular}}
\caption{Main results of our methods on ten different sizes of LLMs across ten datasets. We report the average performance and standard deviation over 5 different seeds for each dataset. The last column represents the average result across the ten datasets.}
\label{Main_results}%
\end{table*}

\begin{table*}[t]
\centering
\scalebox{0.75}{
\begin{tabular}{lcccccc}
\toprule
& \textbf{GPT-2 0.3B} & \textbf{GPT-2 0.8B} & \textbf{GPT-2 1.5B} & \textbf{Cerebras-GPT 2.7B} & \textbf{Cerebras-GPT 6.7B} & \textbf{GPT-3 175B}\\
\midrule
KATE & 66.7 & 69.4 & 67.7 & 71.6 & 77.6 & 82.2\\
\quad + D$^2$Controller & 68.8 & 70.5 & 69.4 & 74.7 & 77.9 & 82.6 \\
\midrule
GlobalE & 59.5 & 67.7 & 69.8 & - & - & -\\
\quad + D$^2$Controller & 61.5 & 68.7 & 71.6 & - & - & -\\
\midrule
Contextual Calibration & 59.5 & 64.2 & 63.9 & 67.2 & 72.5 & 78.9\\
\quad + D$^2$Controller & 60.8 & 66.6 & 65.4 & 68.7 & 73.5 & 80.1\\
\midrule
kNN Prompting & 74.8 & 76.0 & 77.3 & 77.8 & 79.0 & -\\
\quad + D$^2$Controllern & 75.8 & 77.1 & 78.2 & 78.1 & 79.7 & -\\
\bottomrule
\end{tabular}}
\caption{The result of extending D$^2$Controller to other ICL models.}
\label{extend_SL_selection}%
\end{table*}

\begin{table*}[t]
\centering

\scalebox{0.75}{
\begin{tabular}{lcccccc}
\toprule
& \textbf{GPT-2 0.3B} & \textbf{GPT-2 0.8B} & \textbf{GPT-2 1.5B} & \textbf{Cerebras-GPT 2.7B} & \textbf{Cerebras-GPT 6.7B} & \textbf{GPT-3 175B}\\
\midrule
$k_{\max}$-shot setting & 54.1 & 58.7 & 66.0 & 65.4 & 73.0 & 81.4 \\
D$^2$Controller & 54.5 & 62.4 & 67.0 & 68.7 & 74.0 & 82.0 \\
\bottomrule
\end{tabular}}
\caption{The results of D$^2$Controller and using the maximum number of demonstrations.}
\label{compare_with_max}%
\end{table*}

\begin{table*}[t]
\centering
\scalebox{0.95}{
\begin{tabular}{lcccc}
\toprule
& \textbf{GPT-2 1.5B} & \textbf{Cerebras-GPT 2.7B} & \textbf{Cerebras-GPT 6.7B} & \textbf{OPT 13B}\\
\midrule
Default $k$ & 455.49 & 516.87 & 516.87 & 516.87\\
Maximum $k$ & 678.29  & 1345.72 & 1345.72 & 1345.72\\
D$^2$Controller & 603.98 & 885.51 & 1187.37 & 725.89\\
\bottomrule
\end{tabular}}
\caption{The number of tokens used by default $k$, maximum $k$, and D$^2$Controller}
\label{tab:token_num}
\end{table*}

\begin{table*}[t]
\centering
\scalebox{0.78}{
\begin{tabular}{lccccccc}
\toprule
& \textbf{GPT-2 0.3B} & \textbf{GPT-2 0.8B} & \textbf{GPT-2 1.5B} & \textbf{Cerebras-GPT 2.7B} & \textbf{Cerebras-GPT 6.7B} & \textbf{GPT-3 175B}\\
\midrule
Random & 54.1 & 59.2 & 63.5 & 68.0 & 72.9 & 81.3\\
D$^2$Controller-ED & 54.4 & 59.2 & 64.0 & 67.1 & 72.6 & 79.1\\
D$^2$Controller-Cos & \textbf{54.9} & 59.3 & 62.2 & 68.3 & 72.4 & 80.4\\
\midrule
D$^2$Controller & 54.5 & \textbf{62.4} & \textbf{66.9} & \textbf{69.3} & \textbf{74.0} & \textbf{82.0}\\

\bottomrule
\end{tabular}}
\caption{The results of using three other ways to select evaluation examples.}
\label{variants_SL_selection}%
\end{table*}

\begin{table*}[t]
\centering
\scalebox{0.85}{
\begin{tabular}{lcccc}
\toprule
& \textbf{GPT-2 1.5B} & \textbf{Cerebras-GPT 2.7B} & \textbf{Cerebras-GPT 6.7B} & \textbf{OPT 13B}\\
\midrule
Default & 60.0 & 63.0 & 71.7 & 74.5\\
Validation-100 & 64.9 & 68.3 & 72.6 & 75.8\\
Validation-200 & 65.4 & 68.5 & 71.8 & 76.1\\
Validation-300 & 64.9 & 68.3 & 72.6 & 76.4\\
\midrule
D$^2$Controller & \textbf{67.0} & \textbf{69.3} & \textbf{74.0} & \textbf{76.6}\\
\bottomrule
\end{tabular}}
\caption{The results of using validation set sampled from the training dataset.}
\label{tab:use_dev}%
\end{table*}


\section{Experiments}

\subsection{Experimental Setup}\label{sec:ex_setup}

\paragraph{Datasets}
We conduct experiments on ten text classification datasets ranging from sentiment classification to textual entailment, including SST-2~\citep{DBLP:conf/emnlp/SocherPWCMNP13}, SST-5~\citep{DBLP:conf/emnlp/SocherPWCMNP13}, DBPedia~\citep{DBLP:conf/nips/ZhangZL15}, MR~\citep{DBLP:conf/acl/PangL05}, CR~\citep{DBLP:conf/kdd/HuL04}, MPQA~\citep{DBLP:journals/lre/WiebeWC05}, Subj~\citep{DBLP:conf/acl/PangL04}, AGNews~\citep{DBLP:conf/nips/ZhangZL15}, RTE~\citep{DBLP:conf/mlcw/DaganGM05}, and CB~\citep{de2019commitmentbank}. More details of the datasets are provided in Appendix~\ref{sec:Datasets}.


\paragraph{LLMs} To verify the validity of D$^2$Controller, we apply our method to a wide range of LLMs, including three GPT-2 models~\citep{radford2019language} (with 0.3B, 0.8B, and 1.5B parameters), two Cerebras-GPT models~\citep{DBLP:journals/corr/abs-2304-03208} (with 2.7B and 6.7B parameters), two LLAMA models~\citep{DBLP:journals/corr/abs-2302-13971,DBLP:journals/corr/abs-2307-09288} (with 7B parameters), two OPT models~\citep{zhang2022opt} (with 13B and 30B parameters) and GPT-3 175B model~\citep{DBLP:conf/nips/BrownMRSKDNSSAA20}.

\paragraph{Evaluation Metric}
Following~\citep{DBLP:conf/acl/LuBM0S22,xu2023knn}, to control the GPT-3 inference costs~\footnote{It requires the usage of a monetary paid-for API.}, we randomly sample $256$ examples from the validation set for each dataset to evaluate the accuracy and report the average performance and standard deviation over $5$ different seeds.

\paragraph{Implementation Details} In the case of D$^2$Controller, $\mathcal{K}$ is set as $\{1,2,4,8,\cdots,k_{\max}\}$ (See Appendix~\ref{sec:Datasets} for details of $k_{\max}$ of each dataset on different sizes of LLMs). We sample $N_s=5$ groups of in-context examples for $k$-shot setting evaluation on Cerebras-GPT-2.7B model, and set $N_s$ as $25$ on other sizes of LLMs, the reason of which is presented in the Section~\ref{Analysis_and_Discussion}. We implement our method with the PyTorch framework and run the experiments on 8 NVIDIA A100 GPUs. 

\subsection{Base Model and Oracle}\label{sec:baseline_oracle}
We consider the default $k$-shot setting in previous work~\citep{DBLP:conf/acl/LuBM0S22,xu2023knn} as our base model, which is the $4$-shot setting (except the $1$-shot setting for the DBpedia dataset and the $2$-shot setting for the AGNews dataset). In addition, we also provide an \textit{Oracle} to show the upper bound of performance, that is, for each dataset, we iterate all feasible $k$-shot settings on 256 examples (mentioned in Evaluation Metric) and record the highest achievable performance.

\subsection{Main Results}\label{Main_Results}



The main experiment results are shown in Table~\ref{Main_results}, from which we have following findings:

\paragraph{D$^2$Controller is effective for selecting suitable $k$-shot setting for each dataset and is compatible with different LLMs.} In comparison to the base model, D$^2$Controller achieves 4.6\% relative improvements on average across ten datasets, which validates the rationality of dynamically selecting the number of demonstrations\footnote{The values of $k$ chosen by the D$^2$Controller and Oracle are provided in Appendix~\ref{value_k}.}. It is worth mentioning that, in contrast to other LLMs, D$^2$Controller at most obtains 7.0\% and 6.3\% improvements in accuracy for GPT-2-1.5B and Cerebras-GPT-2.7B on ten datasets. These results reveal that our method has good compatibility. Some LLMs exhibit a minor decline in performance on the MPQA, SST-2, and MR datasets. One possible reason is that these datasets have relatively shorter average demonstration lengths (shown in Table \ref{datasets}), leading to encoded semantic representations contain less information. Thus, the similarities measured by IICScore based on these representations are inaccurate. In this case, selecting an appropriate demonstration number for these datasets may be more challenging.

\paragraph{D$^2$Controller achieves near-optimal results at a lower cost.} In most of the LLMs, D$^2$Controller achieves performance levels close to that of the Oracle, aligning with our original research intent. While the Oracle represents the upper bound of performance, it is unfeasible in practice to iterate through all $k$-shot settings on large-scale examples to attain such performance, mainly due to the extensive resource and time demands. Because the Oracle in our paper is obtained on a development set of 256 examples. However, in the real scenario, the number of test examples could be extremely large (maybe thousands of times comparing to the dev set), making it impossible to iterate all $k$-shot settings to decide which one is the best. In contrast, our method achieves good performance with a small number of evaluation examples and effectively controls inference costs. Our method underscores the practical feasibility of striking a balance between performance and resource consumption.

\subsection{Analysis and Discussion}\label{Analysis_and_Discussion}

In this section, we conduct a series of analysis experiments related to D$^2$Controller. It should be noted that the results we report are the average performance of ten datasets.



\paragraph{D$^2$Controller is beneficial to other ICL methods.}We extend our method to some representative ICL methods, \textit{i.e.}, applying the demonstrations number decided by D$^2$Controller to other ICL methods. These methods include a \textit{Demonstration Selection} method \textbf{KATE}~\citep{liu-etal-2022-makes}, a \textit{Demonstration Order} method \textbf{GlobalE}~\citep{DBLP:conf/acl/LuBM0S22}, and two \textit{Calibration-based} method \textbf{Contextual Calibration}~\citep{DBLP:conf/icml/ZhaoWFK021} and \textbf{$k$NN Prompting}~\citep{xu2023knn}. The results are shown in Table~\ref{extend_SL_selection}. 

As we can see, incorporating D$^2$Controller into other ICL methods can obtain competitive performance. Specifically, compared to KATE using the default $k$-shot settings (As mentioned in Section~\ref{sec:baseline_oracle}), KATE + D$^2$Controller obtains 3.1\% improvements in accuracy. Similarly, GlobalE + D$^2$Controller improves the accuracy by up to 2.0\% compared to GlobalE. For Contextual Calibration and $k$NN Prompting, when combined with D$^2$Controller, the accuracy is improved by up to 2.4\% and 1.1\% respectively. For the GPT-3 model, integrating Contextual Calibration with D$^2$Controller enhances accuracy by $1.2\%$. The improvements of these extending methods further confirm the necessity to dynamically decide $k$-shot settings instead of using the default setting as well as indicate that the D$^2$Controller has excellent generalization capabilities. Moreover, the improvements in KATE + D$^2$Controller and GlobalE + D$^2$Controller prove that the number of demonstrations is a key factor in ICL performance along with the selection and ordering of demonstrations.

\begin{figure}[t]
\centering
\scalebox{0.90}{
    \includegraphics[width=\linewidth]{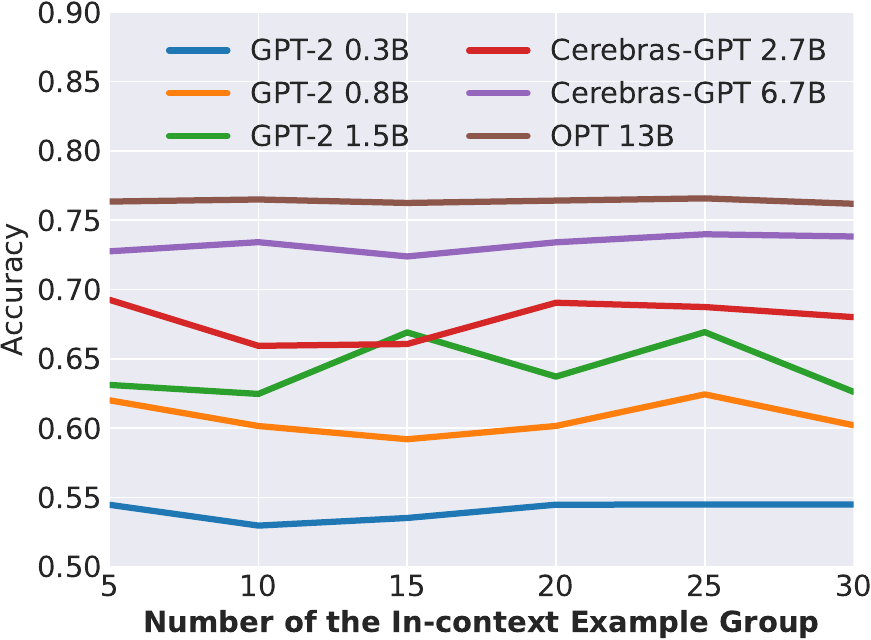}}
  \caption{The impact of the number of in-context example groups $N_s$ on D$^2$Controller.}
    \label{fig:in_context_examples}
\end{figure}

\paragraph{D$^2$Controller can achieve competitive results on a small number of in-context example groups.}
To investigate the effect of the number of in-context example groups $N_s$ on D$^2$Controller, we vary the value of $N_s$ in the range of [5, 30] with a step size of 5. Figure \ref{fig:in_context_examples} shows the average performance of D$^2$Controller with different $N_s$ on ten datasets. Actually, most LLMs can achieve good results at $N_s=5$, and their performance remains stable as the number of in-context example groups increases. For the other LLMs, their performance has an initial upward trend and then flattens out. These observations indicate that D$^2$Controller can select near-optimal $k$-shot settings depending on a small number of in-context example groups. Finally, according to the trend of the curve, we set $N_s$ to 5 in the Cerebras-GPT-2.7B model and set $N_s$ as 25 in other sizes of LLMs. 


\begin{table}[t]
\centering
\scalebox{0.88}{
\begin{tabular}{llc}
\toprule
\textbf{ID} & \textbf{Template} & \textbf{Label Mapping}\\
\midrule
\multirow{2}{*}{Original} & Review: \{Sentence\} &
\multirow{2}{*}{positive/negative}\\
& Sentiment: \{Label\} \\
\midrule
\multirow{2}{*}{T1} & Input: \{Sentence\} &
\multirow{2}{*}{positive/negative}\\
& Prediction: \{Label\} \\
\midrule
\multirow{2}{*}{T2} & Input: \{Sentence\} &
\multirow{2}{*}{good/bad}\\
& Prediction: \{Label\} \\
\bottomrule
\end{tabular}}
\caption{Different templates for SST-2.}
\label{tab:template}
\end{table}

\paragraph{Dynamically selecting $k$ performs better than using the maximum $k$.}

We also compare dynamically selecting the $k$-shot setting (\textit{i.e.}, D$^2$Controller) with using the maximum number of demonstrations (\textit{i.e.}, $k_{\max}$-shot setting). As shown in Table~\ref{compare_with_max}, we observe that D$^2$Controller achieves more competitive results, which agree with our motivation mentioned in Section~\ref{sec:Pilot_Experiments}. Specifically, in contrast to the $k_{\max}$-shot setting, D$^2$Controller achieves a 2.6\% relative improvement across six different sizes of LLMs on ten datasets, indicating that adopting the $k_{\max}$-shot setting for each dataset is not appropriate. 

In addition, we report the average number of tokens used by three methods (default $k$, maximum $k$, and D$^2$Controller) to query LLM. Based on results in Table \ref{tab:token_num}, we can observe that our method uses fewer tokens to achieve better performance compared to maximum $k$. Especially on some LLMs, such as Cerebras-GPT 2.7B and OPT-13B, D$^2$Controller saves almost 30\% and 50\% tokens. Meanwhile, although our method uses more tokens compared to the default $k$, it achieves an average relative improvement of $4.6\%$ on ten datasets.

\paragraph{IICScore is effective in selecting evaluation examples}
Except for IICScore, we also explore other ways to select evaluation examples. As shown in Table~\ref{variants_SL_selection}, \textbf{Random} denotes randomly selecting the same number of examples as that of IICScore. \textbf{D$^2$Controller-ED} and \textbf{D$^2$Controller-Cos} indicate replacing KL divergence in Equation~\ref{eq:IICScore} with Euclidean distance and negative cosine similarity, respectively. It is clear that D$^2$Controller outperforms Random in every LLM, suggesting that the evaluation examples selected by D$^2$Controller are more representative than those of Random to properly reflect the performance of each $k$-shot setting. Comparing D$^2$Controller with the two variants, we can find that both of the variants perform worse than D$^2$Controller on most of the LLMs (except for GPT-2-0.3B), which verifies the superiority of using KL divergence as the distance metric.

Besides, we also randomly sample more examples as a baseline to select $k$. Specifically, we construct three different sizes of validation sets (100, 200, and 300) to select $k$. The results are shown in Table~\ref{tab:use_dev} (note that the results we report are the average performance of ten datasets). Based on these results, we can observe that using more examples does not lead to the optimal choice of $k$, and almost all of the results are inferior to D$^2$Control. This further underscores the effectiveness of using IICScore to select a small number of representative examples.


\begin{figure}[t]
\centering
\scalebox{0.90}{
    \includegraphics[width=\linewidth]{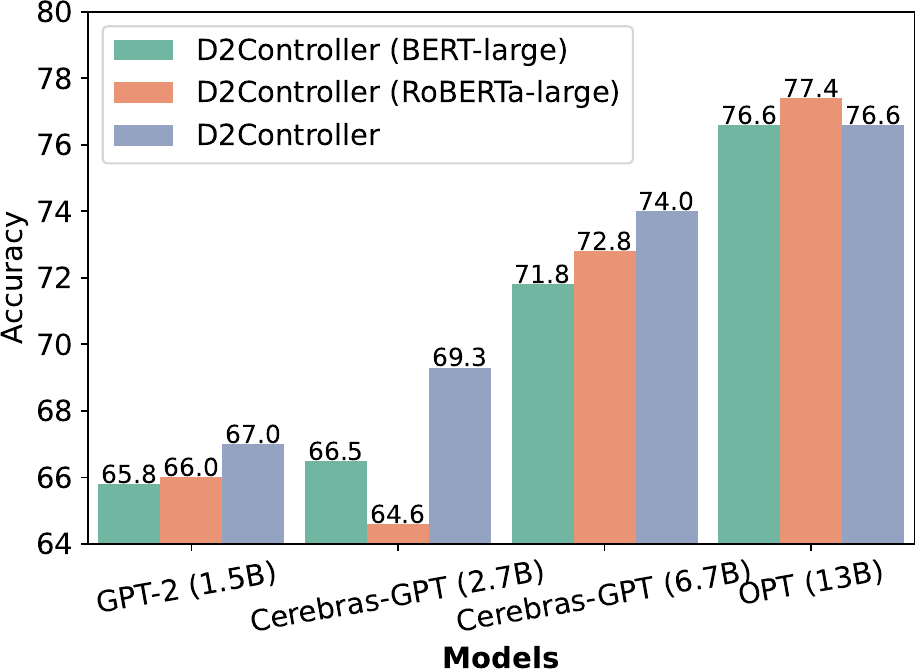}}
  \caption{The results of using BERT-family models as text encoders.}
    \label{tab:bert}%
\end{figure}

\paragraph{Impact of varying prompt templates on the optimal number of demonstrations}
We conduct experiments on the SST-2 dataset with two new templates (T1 and T2) on three GPT-2 family models. The templates and the corresponding selected k-shot settings of them on each LLM are presented in Tables~\ref{tab:template} and Tables~\ref{tab:number}. Based on these results, we can observe that the change of templates could lead to the change of the selected k-shot setting. However, in general, the change in the selected number is minor.

\paragraph{Effect of Different Retrieval Models}
Here, we try another two text encoders (i.e., BERT-large and RoBERTa-large) to obtain sentence representations $\mathbf{x}$. The results are shown in Figure~\ref{tab:bert}.



We observe that D$^2$Controller (BERT-large) and D$^2$Controller (RoBERTa-large) underperform compared to the D$^2$Controller on most of the LLMs, except for OPT 13B. This outcome underscores the advantages of employing GPT-architecture LLMs as text encoders for measuring data similarity in representation space.


\section{Related Work}

With the increase in both model size and training corpus size~\cite{DBLP:conf/naacl/DevlinCLT19,radford2019language,DBLP:conf/nips/BrownMRSKDNSSAA20,DBLP:journals/corr/abs-2204-02311}, large language models (LLMs) have demonstrated a significant capacity for In-Context Learning (ICL). Given that ICL is sensitive to the selection and the order of the demonstrations~\cite{DBLP:conf/acl-deelio/LiuSZDCC22,DBLP:conf/naacl/RubinHB22,DBLP:conf/emnlp/ZhangFT22,DBLP:conf/acl/LuBM0S22,DBLP:journals/corr/abs-2301-11916,DBLP:conf/acl/0003WYK23,DBLP:journals/corr/abs-2305-04320,DBLP:journals/corr/abs-2302-13539,DBLP:conf/acl/LevyBB23,DBLP:conf/iclr/SuKWSWX0OZS023,DBLP:conf/acl/AgrawalZLZG23,DBLP:conf/icml/Ye0F0K23,DBLP:conf/iccv/HeWHLLXS23,DBLP:conf/emnlp/Gupta0023,DBLP:journals/corr/abs-2401-11624,DBLP:conf/emnlp/YangZSLZL23,DBLP:journals/corr/abs-2401-12087}, most of the studies design \textit{Demonstration Selection} methods~\citep{DBLP:conf/acl-deelio/LiuSZDCC22,DBLP:conf/naacl/RubinHB22,DBLP:conf/emnlp/ZhangFT22,DBLP:journals/corr/abs-2206-08082,DBLP:journals/corr/abs-2212-04037,DBLP:conf/acl/SorensenRRSRDKF22} or finding an appropriate \textit{Demonstration Order}~\citep{DBLP:conf/acl/LuBM0S22,wu2022self} to improve the performance of ICL.


\begin{table}[t]
\centering
\scalebox{0.88}{
\begin{tabular}{llccc}
\toprule
\textbf{Models} & \textbf{Original} & \textbf{T1} & \textbf{T2} & \textbf{Optimal}\\
\midrule
GPT-2 0.3B & 16 & 4 & 8 & 16\\
GPT-2 0.8B & 16 & 16 & 16 & 16\\
GPT-2 1.5B & 16 & 16 & 16 & 16\\
\bottomrule
\end{tabular}}
\caption{The value of $k$ selected by GPT-2 family models under different templates.}
\label{tab:number}
\end{table}


However, there are few studies related to the impact of the number of demonstrations within a limited input length on ICL performance. The closest work to ours is~\cite{xu2023knn}, which proposes a method that utilizes an unlimited number of training examples for model calibration, while our research focuses on how to select an appropriate number of demonstrations for each dataset when the input length is restricted. Therefore, the two methods have different starting points.

\section{Conclusion}
\label{sec:bibtex}


In this paper, we conduct an in-depth analysis of the impact of the number of demonstrations on ICL performance. Surprisingly, we discover that the number of demonstrations does not always exhibit a positive correlation with model performance. Based on this, we develop D$^2$Controller that can dynamically select the number of demonstrations. The results show our method achieves an average of 4.6\% relative improvement across ten datasets on ten different sizes of LLMs. 


\section*{Limitations}

The current research suffers from two limitations: (1) Due to budget constraints and insufficient GPU memory, we are unable to conduct experiments on larger-scale language models; (2) Our method does not guarantee the selection of the optimal value of $k$ for each dataset. As we mentioned in section~\ref{Main_Results}, some LLMs exhibit a minor decline in performance on the MPQA, SST-2, and MR datasets compared to the default setting. This indicates the need for future research to further refine the selection of $k$ to approach its optimal value.



\bibliography{tacl2021}
\bibliographystyle{acl_natbib}


\onecolumn

\appendix

\section{Detail for Demonstration and Label Space}\label{sec:Template}

As depicted in Table~\ref{table:template}, we provide detailed information on the Demonstration, mapping token space, and label space for different tasks.

\begin{table*}[htbp]
\centering
\scalebox{0.8}{
\begin{tabular}{llll}
\toprule
Dataset & Demonstration & Mapping Token Space $\mathcal{V}$ & Label Space $\mathcal{Y}$\\
\midrule
SST-2 & \makecell[l]{Review: the greatest musicians. \\ Sentiment: Positive} & positive/negative & positive/negative\\
\midrule
SST-5 & \makecell[l]{Review: it 's a very valuable film ... \\ Sentiment: great} & \makecell[l]{terrible/bad/okay \\ /good/great} & \makecell[l]{very positive/positive \\ /neutral/negative \\ /very negative}\\
\midrule
DBPedia & \makecell[l]{input: Monte Vermenone is a mountain\\ of Marche Italy. \\ type: nature} & \makecell[l]{company/school/artist/ \\ athlete/politics/book/ \\ building/nature/village/
\\ animal/plant/album/ \\ film/transportation} & \makecell[l]{company/school/artist/ \\ athlete/politics/book/ \\ building/nature/village/
\\ animal/plant/album/ \\ film/transportation}\\
\midrule
MR & \makecell[l]{Review: a dreary movie . \\ Sentiment: negative} & positive/negative & positive/negative\\
\midrule
CR & \makecell[l]{Review: i am bored with the silver look . \\ Sentiment: negative} & positive/negative & positive/negative\\
\midrule
MPQA & \makecell[l]{Review: is also the most risky \\ Sentiment: negative} & positive/negative & positive/negative\\
\midrule
Subj & \makecell[l]{Input: presents a most persuasive \\vision of hell on earth . \\ Type: subjective} & subjective/objective & subjective/objective\\
\midrule
AGNews & \makecell[l]{input: Historic Turkey-EU deal welcomed. The \\European Union's decision to hold entry talks with\\ Turkey receives a widespread welcome.\\ type: world} & \makecell[l]{world/sports/business \\ /technology} & \makecell[l]{world/sports/business \\ /technology}\\
\midrule
RTE & \makecell[l]{premise: Oil prices fall back as Yukos oil threat lifted \\ hypothesis: Oil prices rise. \\ prediction: not\_entailment} & true/false & entailment/not\_entailment\\
\midrule
CB & \makecell[l]{premise: ``Clever''. Klug means ``clever''. Would \\you say that Abie was clever? \\ hypothesis: Abie was clever \\ prediction: neutral} & true/false/neither & \makecell[l]{entailment/contradiction/ \\ neutral}\\
\bottomrule
\end{tabular}}
\caption{Demonstration, mapping token space, and label space for different tasks.}
\label{table:template}%
\end{table*}

\section{Detail for Datasets and Max Shots}\label{sec:Datasets}

As shown in Table~\ref{datasets}, we present detailed information for ten datasets. Besides, as we mentioned in section~\ref{In_Context_Learning}, for each dataset, the input prompt $P$ consists of different numbers of demonstrations and a test instance. The maximum shot number, \textit{i.e.}, $k_{\max}$ is calculated as follows:
\begin{equation}
\text{Upper}_{bound} = \frac{\text{Max}_{input}-\text{Max}_{test}}{\text{Avg}_{template} * \text{Numbers}_{classes}},
\end{equation}
\begin{equation}
k_{\max} = \max\,2^i\leq\text{Upper}_{bound},\quad i=0,1,2,\cdots
\end{equation}
where $\text{Upper}_{bound}$ is the Upper-bound of shots that can be accommodated by GPT-2, Cerebras-GPT, OPT or GPT-3, $\text{Max}_{input}$ indicates the maximum input length of different sizes of LLMs, i.e., GPT-2 (1024 tokens), Cerebras-GPT-2.7B (2048 tokens), Cerebras-GPT-6.7B (2048 tokens), OPT-13B (2048 tokens), OPT-30B (2048 tokens), GPT-3 175B (2048 tokens), $\text{Max}_{test}$ denotes the max length of test input, $\text{Avg}_{template}$ means the average length of each demonstration, and $\text{Numbers}_{classes}$ indicates the numbers of classes for each task, \textit{i.e.}, $|\mathcal{C}|$. To narrow down the search scope, we set the value range of Max Shots to $\{1, 2, 4, 8, 16, 32, 64, \cdots\}$. Thus, for each dataset, the max shots we choose should be below the upper bound and closest to it. For example, the Upper-bound (1024 tokens) of the SST-2 dataset is 25, so the max shot we need to select is 16; the Upper-bound (1024 tokens) of the MPQA dataset is 48, so the max shot we need to select is 32. It should be noted that while the Upper-bound (1024 tokens) of the CB dataset is 2, for a fair comparison with other methods, we set the max shot to 4. This decision was made because previous methods used 4-shots for the CB dataset~\citep{DBLP:conf/acl/LuBM0S22}.

\begin{table*}[t]
\centering
\scalebox{0.68}{
\begin{tabular}{lccccccc}
\toprule
Dataset & \makecell[c]{Number of \\ Classes} & \makecell[c]{Avg. Length \\ of Demonstration} & \makecell[c]{Max Length of \\ Test Input} & \makecell[c]{Upper-bound \\ (1024 tokens)} & \makecell[c]{Max Shots \\ (1024 tokens)} & \makecell[c]{Upper-bound \\ (2048 tokens)} & \makecell[c]{Max Shots \\ (2048 tokens)}\\
\midrule
SST-2~\citep{DBLP:conf/emnlp/SocherPWCMNP13} & 2 & 19.1 & 55 & 25 & 16 & 52 & 32\\
SST-5~\citep{DBLP:conf/emnlp/SocherPWCMNP13} & 5 & 29.7 & 60 & 6 & 4 & 13 & 8\\
DBPedia~\citep{DBLP:conf/nips/ZhangZL15} & 14 & 71.6 & 161 & 1 & 1 & 1 & 1 \\
MR~\citep{DBLP:conf/acl/PangL05} & 2 & 32.7 & 66 & 14 & 8 & 30 & 16 \\
CR~\citep{DBLP:conf/kdd/HuL04} & 2 & 29.0 & 99 & 15 & 8 & 33 & 32 \\
MPQA~\citep{DBLP:journals/lre/WiebeWC05} & 2 & 10.4 & 19 & 48 & 32 & 97 & 64 \\
Subj~\citep{DBLP:conf/acl/PangL04} & 2 & 34.9 & 91 & 13 & 8 & 28 & 16\\
AGNews~\citep{DBLP:conf/nips/ZhangZL15} & 4 & 59.5 & 167 & 3 & 2 & 7 & 4\\
RTE~\citep{DBLP:conf/mlcw/DaganGM05} & 2 & 79.7 & 256 & 4 & 4 & 11 & 8 \\
CB~\citep{de2019commitmentbank} & 3 & 90.8 & 278 & 2 & 4 & 6 & 4 \\
\bottomrule
\end{tabular}}
\caption{Statistics of evaluation datasets, the average length of each demonstration, and the max length of test input are calculated based on sentence-piece length.}
\label{datasets}%
\end{table*}


\section{Additional Pilot Experiments}\label{Additional_Results}


\begin{figure*}[t]
\centering
\begin{minipage}{0.45\linewidth}
    \centering
    \includegraphics[width=\linewidth]{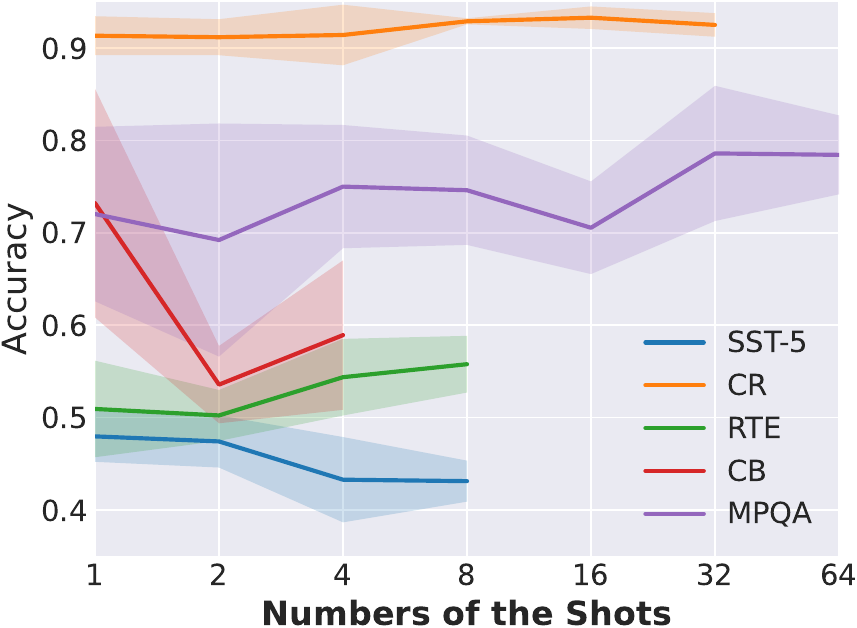}
    \caption{Effect of the number of demonstrations on OPT-13B across five datasets.}
    \label{fig:pilot_13B}
\end{minipage}
\hspace{11pt}
\begin{minipage}{0.45\linewidth}
    \centering
    \includegraphics[width=\linewidth]{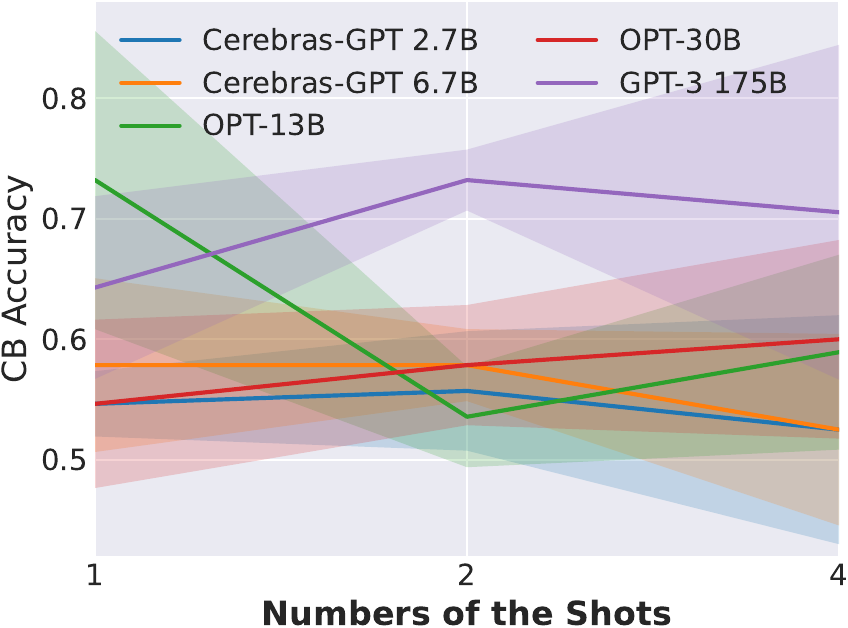}
    \caption{The accuracy of five different sizes of LLMs on the CB dataset.}
    \label{fig:pilot_cb}
\end{minipage}
\end{figure*}

\begin{table}[t]
\centering

\scalebox{0.95}{
\begin{tabular}{lccccc}
\toprule
\textbf{GPT-4} & \textbf{SST-5} & \textbf{CR} & \textbf{MPQA} & \textbf{RTE} & \textbf{CB} \\
\midrule
$1$-shot setting & 45.3$_{4.4}$ & 83.7$_{1.3}$ & 67.4$_{1.0}$ & 82.7$_{3.0}$ & 89.3$_{1.8}$ \\
\midrule
Default setting & 45.7$_{5.0}$ & 92.2$_{2.2}$ & 83.8$_{0.3}$ & 89.1$_{1.4}$ & 83.9$_{2.5}$\\
\midrule
$k_{\max}$-shot setting & 43.6$_{0.8}$ & 95.9$_{0.3}$ & 90.2$_{1.1}$ & 88.7$_{0.6}$ & 82.7$_{1.0}$\\
\bottomrule
\end{tabular}}
\caption{The results of using the $1$-shot setting, default setting, and the $k_{\max}$-shot setting on GPT-4.}
\label{tab:GPT-4}%
\end{table}

Here, we present more results to support our arguments. Among them, Figure~\ref{fig:pilot_13B} shows the performance curves of five datasets on the OPT-13B model. Figure~\ref{fig:pilot_cb} shows performance curves of CB dataset on five different sizes of LLMs. Besides, we also conduct experiments with the GPT-4 model on five datasets, the results are shown in Table~\ref{tab:GPT-4}. For the GPT-4 model, due to budgetary constraints, we use five different seeds to test model's performance in the 1-shot setting, the default setting (4-shot), and $k_{max}$-shot setting. Note that the maximum input length of the GPT-4 we use is 8192 tokens, so the maximum shot number for SST-5, CR, MPQA, RTE, and CB is 32, 128, 256, 32, and 16.

\paragraph{Increasing the number of demonstrations does not necessarily improve the model performance.} In Figure~\ref{fig:pilot_13B}, when changing from $1$-shot setting to $k_{\max}$-shot setting, we can observe that the accuracy of the OPT-13B model improves on the RTE and MPQA datasets while declines on the SST5 and CB datasets. Besides, as shown in Figure~\ref{fig:pilot_cb}, when changing from $1$-shot setting to $4$-shot setting, the accuracy of the CB dataset initially declines and then increases on the OPT-13B model, while it first rises and then goes down on the GPT-3-175B model. Even for stronger LLM such as GPT-4, as observed from the overall trend in Table~\ref{tab:GPT-4}, when the input increases from a $1$-shot setting to $k_{max}$- shot setting, the accuracy improves on the CR, MPQA, and RTE datasets while declines on the SST-5 and CB datasets. These observations suggests that the inclusion of more demonstrations does not guarantee improved performance.

\paragraph{The optimal $k$-shot setting differs depending on specific datasets and models.} From Figure~\ref{fig:pilot_cb}, we can find that the optimal $k$-shot settings for the same dataset on different models can be different: $1$-shot setting for the OPT-13B model, $2$-shot setting for the Cerebras-GPT 2.7B, Cerebras-GPT 6.7B and GPT-3 175B models, $4$-shot setting for the OPT-30B model. Likewise, from Figure~\ref{fig:pilot_13B}, we can tell that the optimal $k$-shot settings for the same model on different datasets also can be different: $1$-shot setting for the SST5 and CB datasets, $8$-shot setting for the RTE dataset, $16$-shot setting for the CR dataset, $32$-shot setting for the MPQA dataset. These observations suggests that the optimal number of demonstrations may differ depending on the specific dataset and model.

\section{Algorithm details}\label{Algorithm_details}
The details of the Dynamic Demonstrations Controller are presented in Algorithm~\ref{alg:algorithm1}.

\begin{algorithm}[h]
\begin{small}
	\caption{Dynamic Demonstrations Controller.}
	\label{alg:algorithm1}
	\KwIn{The training set: $\mathcal{D}$; The number of in-context example groups: $N_s$; The feasible $k$ set: $\mathcal{K}$; The set of Classes: $\mathcal{C}$;The LLM: $\theta$.}
	\KwOut{The selected $k$: $\hat{k}$.}
	\BlankLine
   
    \For{$k$ in $\mathcal{K}$}{
        Sampling $N_s$ groups of in-context examples and remove them  from $\mathcal{D}$. The rest is $\mathcal{D}'$.

        \tcp{Initializing the set of evaluation examples.}
        $\mathcal{T}^k\leftarrow\emptyset$ \qquad
        
        \For{$i$ in $1,2,\cdots,N_s$}{
            \For{$c$ in $\mathcal{C}$}{
                \tcp{Computing the IICScore for each candidate example in $\mathcal{D}'$.}
                $\tilde{e}^{c}_{\mathcal{E}^k_i}\leftarrow\underset{e^c_j\in\mathcal{D}'}{\operatorname{\arg\max}}\quad\text{IICScore}(e^c_j,\mathcal{E}^k_i)$\qquad 
                
                $\mathcal{T}^k\leftarrow\mathcal{T}^k\cup \tilde{e}^{c}_{\mathcal{E}^k_i}$
            }
        }
        
        {{$\text{Acc}\leftarrow0$}}
        
        \For{$i$ in $1,2,\cdots,N_s$}{
            $\text{Acc}\leftarrow\text{Acc}+\frac{1}{|\mathcal{T}^k|}\sum_{j=1}^{|\mathcal{T}^k|}\mathbb{I}(\hat{y}_{j,\mathcal{E}_i^k}=y_{j})$
        }
        
        $\text{Acc}_k\leftarrow\frac{1}{N_s}\text{Acc}$
        
    }
    $\hat{k}\leftarrow\underset{k\in\mathcal{K}}{\operatorname{\arg\max}}\quad\text{Acc}_k$
    
    \Return $\hat{k}$
\end{small}
\end{algorithm}

\section{The Value of $k$}\label{value_k}

In Table~\ref{k_value}, we show the values of $k$ chosen by the D$^2$Controller and \textit{Oracle}.

\section{The Running Times for D$^2$Controller}
In this section, we provide running times for three different sizes of LLMs during the \textbf{Evaluation Examples Selection} and \textbf{Accuracy-based Evaluation} stages in Table \ref{tab:running_time}, respectively.




\begin{table*}[htbp]
\centering
\scalebox{0.85}{
\begin{tabular}{llcccccccccc}
\toprule
& & \textbf{SST-2} & \textbf{SST-5} & \textbf{DBPedia} & \textbf{MR} & \textbf{CR} & \textbf{MPQA} & \textbf{Subj} & \textbf{AGNews} & \textbf{RTE} & \textbf{CB}\\
\midrule




\multirow{3}{*}{\makecell[l]{GPT-2\\0.3B}} & Default & 4 & 4 & 1 & 4 & 4 & 4 & 4 & 2 & 4 & 4 \\

& D$^2$Controller & 16 & 1 & 1 & 8 & 1 & 32 & 2 & 2 & 2 & 4\\

& Oracle & 16 & 1 & 1 & 1 & 1 & 16 & 8 & 1 & 4 & 2\\

\midrule
\multirow{3}{*}{\makecell[l]{GPT-2\\0.8B}} & Default & 4 & 4 & 1 & 4 & 4 & 4 & 4 & 2 & 4 & 4 \\

& D$^2$Controller & 16 & 2 & 1 & 4 & 4 & 32 & 8 & 2 & 4 & 2\\

& Oracle & 4 & 1 & 1 & 4 & 4 & 16 & 8 & 2 & 2 & 1\\

\midrule
\multirow{3}{*}{\makecell[l]{GPT-2\\1.5B}} & Default & 4 & 4 & 1 & 4 & 4 & 4 & 4 & 2 & 4 & 4 \\

& D$^2$Controller & 16 & 4 & 1 & 8 & 8 & 16 & 8 & 2 & 2 & 4\\

& Oracle & 16 & 1 & 1 & 8 & 8 & 16 & 8 & 1 & 2 & 2\\



\midrule
\multirow{3}{*}{\makecell[l]{Cerebras-GPT\\2.7B}} & Default & 4 & 4 & 1 & 4 & 4 & 4 & 4 & 2 & 4 & 4 \\
& D$^2$Controller & 32 & 8 & 1 & 16 & 1 & 32 & 16 & 1 & 4 & 1\\
& Oracle & 8 & 8 & 1 & 16 & 1 & 64 & 16 & 1 & 2 & 2\\

\midrule
\multirow{3}{*}{\makecell[l]{Cerebras-GPT\\6.7B}} & Default & 4 & 4 & 1 & 4 & 4 & 4 & 4 & 2 & 4 & 4 \\

& D$^2$Controller & 32 & 2 & 1 & 8 & 32 & 64 & 16 & 4 & 8 & 1\\

& Oracle & 1 & 1 & 1 & 4 & 16 & 64 & 16 & 4 & 2 & 2\\

\midrule
\multirow{3}{*}{\makecell[l]{LLAMA\\7B}} & Default & 4 & 4 & 1 & 4 & 4 & 4 & 4 & 2 & 4 & 4 \\
& D$^2$Controller & 4 & 4 & 1 & 4 & 4 & 4 & 1 & 4 & 4 & 2\\
& Oracle & 1 & 1 & 1 & 16 & 32 & 32 & 4 & 2 & 4 & 2\\
\midrule
\multirow{3}{*}{\makecell[l]{LLAMA-2\\7B}} & Default & 4 & 4 & 1 & 4 & 4 & 4 & 4 & 2 & 4 & 4 \\
& D$^2$Controller & 1 & 8 & 1 & 8 & 16 & 16 & 1 & 2 & 4 & 1\\
& Oracle & 32 & 8 & 1 & 8 & 32 & 64 & 4 & 4 & 8 & 4\\
\midrule
\multirow{3}{*}{\makecell[l]{OPT\\13B}} & Default & 4 & 4 & 1 & 4 & 4 & 4 & 4 & 2 & 4 & 4 \\

& D$^2$Controller & 16 & 4 & 1 & 4 & 1 & 1 & 16 & 4 & 8 & 4\\

& Oracle & 1 & 1 & 1 & 1 & 16 & 32 & 16 & 4 & 8 & 1\\

\midrule
\multirow{3}{*}{\makecell[l]{OPT\\30B}} & Default & 4 & 4 & 1 & 4 & 4 & 4 & 4 & 2 & 4 & 4 \\
& D$^2$Controller & 4 & 8 & 1 & 16 & 2 & 64 & 16 & 4 & 8 & 4\\
& Oracle & 2 & 1 & 1 & 16 & 16 & 64 & 16 & 2 & 4 & 4\\

\midrule
\multirow{3}{*}{\makecell[l]{GPT-3\\175B}} & Default & 4 & 4 & 1 & 4 & 4 & 4 & 4 & 2 & 4 & 4 \\
& D$^2$Controller & 4 & 8 & 1 & 16 & 1 & 4 & 16 & 2 & 2 & 2\\

& Oracle & 2 & 8 & 1 & 8 & 2 & 32 & 16 & 2 & 8 & 2\\

\bottomrule
\end{tabular}}
\caption{The values of $k$ chosen by the D$^2$Controller and \textit{Oracle}.}
\label{k_value}%
\end{table*}


\begin{table*}[htbp]
\centering
\scalebox{0.85}{
\begin{tabular}{lccccccccc}
\toprule
& \textbf{SST-2} & \textbf{SST-5} & \textbf{MR} & \textbf{CR} & \textbf{MPQA} & \textbf{Subj} & \textbf{AGNews} & \textbf{RTE} & \textbf{CB} \\
\midrule
\multicolumn{10}{c}{\textbf{GPT-2 1.5B}}\\
\midrule
Evaluation Examples Selection & 1364 s & 313 s & 158 s & 31 s & 189 s & 140 s & 1900 s & 36 s & 10 s \\
Accuracy-based Evaluation & 915 s & 1978 s & 753 s & 654 s & 1112 s & 806 s & 1105 s & 904 s & 1987 s\\
\midrule
\multicolumn{10}{c}{\textbf{Cerebras-GPT 2.7B}}\\
\midrule
Evaluation Examples Selection & 1662 s & 356 s & 183 s & 22 s & 197 s & 158 s & 2943 s & 47 s & 10 s \\
Accuracy-based Evaluation & 2360 s & 5386 s & 1946 s & 3654 s & 2778 s & 2096 s & 3242 s & 2419 s & 2694 s\\
\midrule
\multicolumn{10}{c}{\textbf{ Cerebras-GPT 6.7B}}\\
\midrule
Evaluation Examples Selection & 1685 s & 405 s & 189 s & 21 s & 188 s & 170 s & 2825 s & 45 s & 10 s\\
Accuracy-based Evaluation & 4832 s & 10725 s & 3942 s & 7076 s & 5558 s & 4223 s & 6432 s & 4773 s & 5376 s\\

\bottomrule
\end{tabular}}
\caption{The running times for three different sizes of LLMs during the \textbf{Evaluation Examples Selection} and \textbf{Accuracy-based Evaluation} stages.}
\label{tab:running_time}
\end{table*}

\end{document}